\documentclass[fleqn,10pt]{wlscirep}
\usepackage[utf8]{inputenc}
\usepackage[T1]{fontenc}
\usepackage{graphicx}
\usepackage{amsmath}
\title{Emergence of Machine Language: Towards Symbolic Intelligence with Neural Networks}

\author[1,2,3]{Yuqi Wang}
\author[2,3,*]{Xu-Yao Zhang}
\author[2,3]{Cheng-Lin Liu}
\author[1,2,3,*]{Zhaoxiang Zhang}
\affil[1]{Center for Research on Intelligent Perception and Computing, Institute of Automation, Chinese Academy of Sciences, 100190 Beijing, China.}
\affil[2]{National Laboratory of Pattern Recognition, Institute of Automation, Chinese Academy of Sciences, 100190 Beijing, China.}
\affil[3]{School of Artificial Intelligence, University of Chinese Academy of Sciences, 100049 Beijing, China.}
\affil[*]{Corresponding: zhaoxiang.zhang@ia.ac.cn, xyz@nlpr.ia.ac.cn}

\begin{abstract}
Representation is a core issue in artificial intelligence. Humans use \emph{discrete} language to communicate and learn from each other, while machines use \emph{continuous} features (like vector, matrix, or tensor in deep neural networks) to represent cognitive patterns. Discrete symbols are low-dimensional, decoupled, and have strong reasoning ability, while continuous features are high-dimensional, coupled, and have incredible abstracting capabilities. In recent years, deep learning has developed the idea of continuous representation to the extreme, using millions of parameters to achieve high accuracies. Although this is reasonable from the statistical perspective, it has other major problems like lacking interpretability, poor generalization, and is easy to be attacked. Since both paradigms have strengths and weaknesses, a better choice is to seek reconciliation. In this paper, we make an initial attempt towards this direction. Specifically, we propose to combine symbolism and connectionism principles by using neural networks to derive a discrete representation. This process is highly similar to \emph{human language}, which is a natural combination of discrete symbols and neural systems, where the brain processes continuous signals and represents intelligence via discrete language. To mimic this functionality, we denote our approach as \emph{machine language}. By designing an interactive environment and task, we demonstrated that machines could generate a spontaneous, flexible, and semantic language through cooperation. Moreover, through experiments we show that discrete language representation has several advantages compared with continuous feature representation, from the aspects of interpretability, generalization, and robustness.

\end{abstract}
\begin{document}

\flushbottom
\maketitle

\thispagestyle{empty}

\section*{1 \quad Introduction}

\qquad What is a better representation of cognitive patterns? There are many explorations in the development of artificial intelligence. Early \textbf{symbolism} \cite{nilsson1998artificial} used logical symbols to represent patterns which dominated the field in the 20th century. Discrete symbols are a good embodiment of logical reasoning and naturally have good interpretation ability. On the other side, inspired by neuroscience and brain researches~\cite{bear2020neuroscience}, \textbf{connectionism}~\cite{fodor1988connectionism} used artificial neural networks to learn continuous representation from a large amount of data, which can outperform symbolic intelligence significantly from the perspective of accuracy. In recent years, deep learning or deep neural networks have become the dominant method in artificial intelligence. We can now automatically recognize thousands of objects in natural images~\cite{russakovsky2015imagenet}, generate vivid image captions~\cite{xu2015show,vinyals2015show}, and answer complex questions~\cite{antol2015vqa} about scenes. Although connectionism has achieved great success nowadays, we need to consider its limitations and weaknesses, in order to further promote this field.

Current deep neural network uses high-dimensional vectors, matrices, and tensors to represent cognitive patterns, by combining millions of neurons to obtain powerful representation capabilities. Therefore, there is no surprise that it could achieve high accuracies in different tasks. However, the drawbacks come from other perspectives such as: lacking interpretability, poor generalization, and poor robustness. On the contrary, these properties are actually inherent functions of humans, which use discrete language to represent their patterns. The development of human language can provide some enlightenment to improve our models and algorithms. The Analects~\cite{waley2005analects} of thousands of years ago can still be read and understood by people today. A few short words can contain a wealth of truth. Low-dimensional decoupled symbols are more explanatory. Languages with powerful abstraction capabilities make generalization performance better. The inherent structure and logic of the language also make it robust in complex situations.

Language is a unique hallmark of the human species. Babies hear language to understand the world, students adopt language to acquire knowledge, and adults use language to communicate and cooperate. Language plays an indispensable role in human intelligence. However, \textbf{the emergence of human language is still a mystery} \cite{chomsky1986knowledge, macwhinney1999emergence}. The growth of human language can be summarized into three stages: (1) emergence: from the monosyllable sound of the early primitive people to the early hieroglyphics that formed the tribe, (2) development: the invention of the written language for better recording and communication, and (3) evolution: generating a unique language with structure and syntax when combined with particular culture and custom. Human language was not created overnight but experienced a long-term development (hundreds or thousands of years). Social labor and cultural exchanges have a strong influence on language emergence, development, and evolution. It is very complicated to explore these issues from the current language, because it is not only the expression of semantics but also includes syntactic structure, rhetoric, and cultural customs.

\begin{figure}[t]
  \centering
  \includegraphics[scale=0.35]{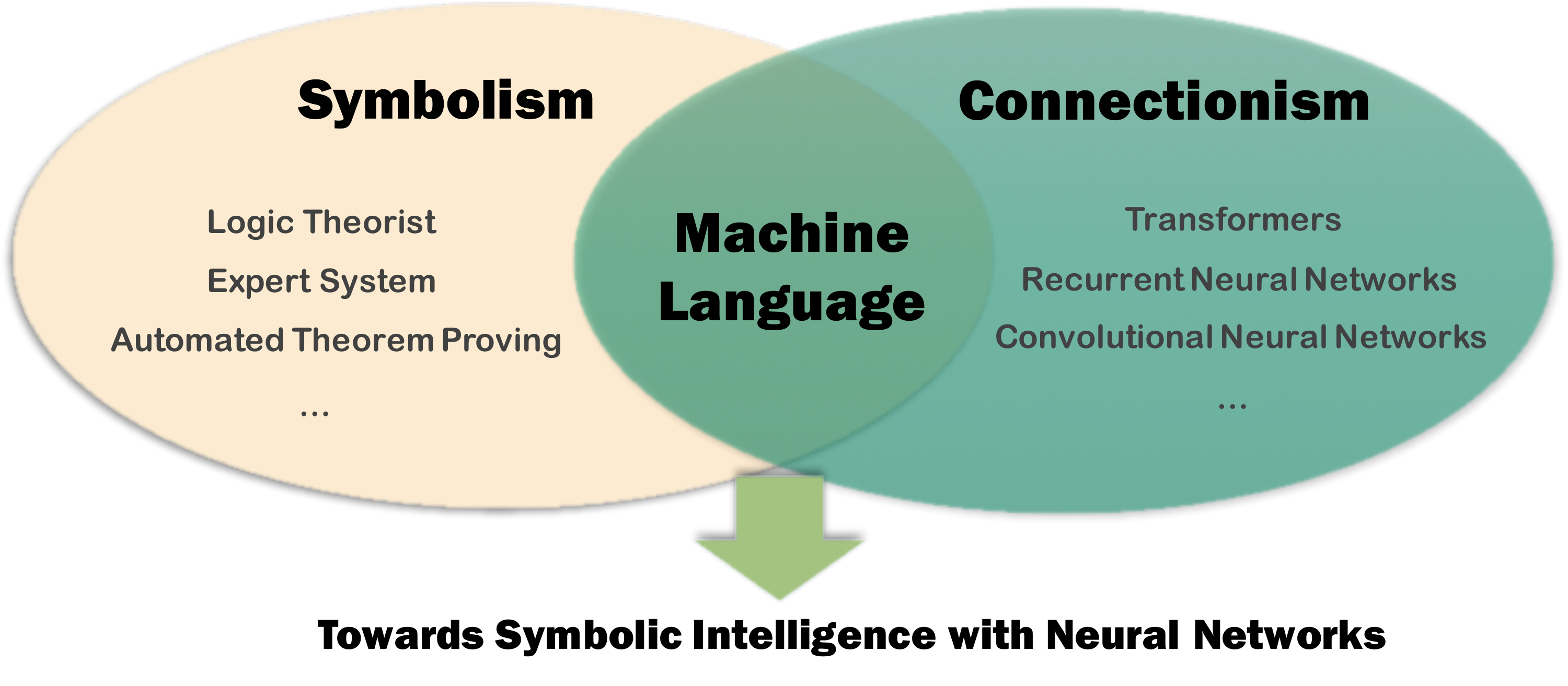}
  \caption{Machine language can be viewed as a combination of symbolism and connectionism.}
  \label{fig:fig1}
\end{figure}

In this paper, inspired from the advantages of human language and the weaknesses of current intelligent models, we propose a new method by combining deep neural networks with symbolic intelligence, to derive a new representation which we call machine language. This is an initial but important attempt at the emergence of a non-existent language among machines. Compared with human language, we focus more on the semantics of the language and put aside syntax and structure. In other words, we care about the emergence other than the development and evolution of the language. As shown in Figure \ref{fig:fig1}, our approach can be viewed as a combination of symbolism and connectionism, which could therefore inherit the advantages from both of them. Besides showing the emergence of machine language, we also verified its functionality by comparing discrete language with the continuous feature from three aspects of interpretability, generalization, and robustness, on diverse datasets and tasks.

\section*{2 \quad Emergence of Machine Language}

\subsection*{2.1 \quad Basic Nature of Machine Language}\label{sec:nature}
\qquad Human language is a complex system that is developing constantly according to the changes from social development and cultural evolution. As a preliminary attempt, machine language is impossible to meet all the properties of human language. Therefore, we focus more on the emergence of machine language from the perspective of semantics. To achieve this goal, we need to figure out what makes a sequence of discrete symbols become a language. Inspired by the characteristics of human language \cite{fromkin2018introduction}, we propose three natures that machine language needs to meet. Just like the early languages in tribes, although being simple in form and grammar, they should still have some basic natures that are crucial to becoming a language.

\begin{itemize}
    \item \textbf{Spontaneous:} The emergence of machine language should be spontaneous. The prior knowledge from human language and data annotation should not be considered. In other words, the process of language emergence should be unsupervised or self-supervised, just like primitive men improving the representation of their language in practice.
    \item \textbf{Flexible:} The form of language should be flexible, i.e., it should be a discrete symbol sequence with variable length. This is because the descriptions of the same objects differ from individual to individual, which can be long or short, concisely or elaborately. Moreover, language should have different vocabularies, for example, English has 26 characters while Chinese has more than thousands of characters.
    \item \textbf{Semantic:} Through the permutation and combination of basic symbols, a language should contain semantics. It should be communicable and understandable between machines, and it can be used to complete specific tasks, like describing something or guiding others to do something.
\end{itemize}

\begin{figure}[t]
  \centering
  \includegraphics[scale=0.5]{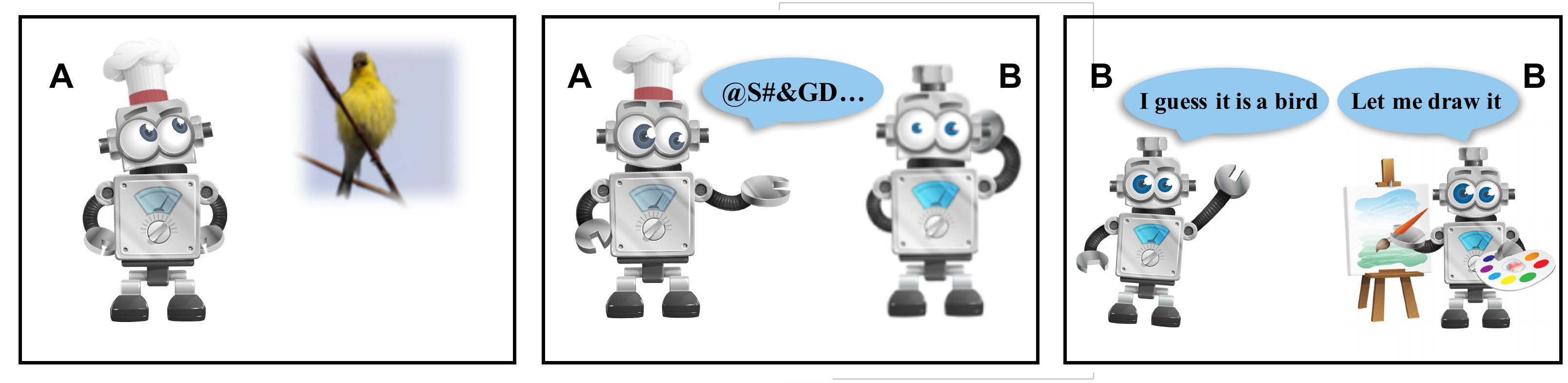}
  \caption{Emergence of language through a game of Speak, Guess and Draw. From left to right: Agent A sees a random image; A tries to use his own language (to be learned) to tell a friend B what he saw; B needs to understand the language and guess what A is talking about and meanwhile draw the image according to the description of A.}
  \label{fig:fig2}
\end{figure}

\subsection*{2.2 \quad From Agent Cooperation to Language}
\label{sec:Game}
\qquad To reach the above three objectives, a basic idea is using the cooperation among multiple machine agents to learn a language automatically. This process should involve multiple agents solving multiple tasks in some complex environments. To simulate this process, we start from the simplest setting of generating a language under a two-agent game. As shown in Figure~\ref{fig:fig2}, two agents are playing together: Speaker $A$ and Listener $B$. The emergence of language can be divided into three stages, which correspond to the three scenes above: (1) Perception: $A$ observes a target image; (2) Communication: $A$ and $B$ are communicating through a sequence of symbols; and (3) Cooperation: They need to solve tasks based on communication. Specifically, Agent $A$ sees the target image $bird$ sitting on the tree, and $A$ tries to tell his friend $B$ what he saw in language. However, because the language is novel, $B$ sounds slightly confused at first. Then they need to solve multi-tasks based on the emergent language. Firstly, $B$ needs to understand the language in the communication and guess what $A$ is talking about. At the same time, $B$ needs to draw the original target to reflect whether he really understands the language. They would get rewards when playing well and punishments otherwise. This simple game is called \textbf{SGD (Speak, Guess, and Draw)} in this paper. The task of guessing represents a macroscopic understanding, while the task of the painting represents a microscopic reconstruction of linguistic description. The proposed game $G$ can be characterized by a tuple:
\begin{equation}
G = <D, V, R, A_s, A_l, M>, \qquad B = <T,*,...,*>.
\end{equation}

$D$ is the set of all images. $V$ is the vocabulary, which limits the symbols that the agent can use, for example, $26$ characters in English. $R$ is the length range of the sequence. We randomly give a length $r \in R$ for each turn. $A_s$ is the speaker agent. The speaker observes the target image and outputs a variable-length sequence $M=(m_1,m_2,...,m_r)$ with length $r$. $A_l$ is the listener agent. The listener hears the machine language and decodes the information to solve two tasks: one is to guess the target with some distractors, the other is to draw the target according to the information. They will get a reward when $A_l$ guesses right and draws similarly and a punishment otherwise. In the beginning, $A_s$ and $A_l$ are like babies, knowing nothing about how to communicate, but by playing the game together, they gradually learn how to describe and communicate, leading to the emergence of machine language $M$. We pick a batch of pictures $B$ at a time and choose a target image $T$ randomly.

\begin{figure}[t]
  \centering
  \includegraphics[scale=0.5]{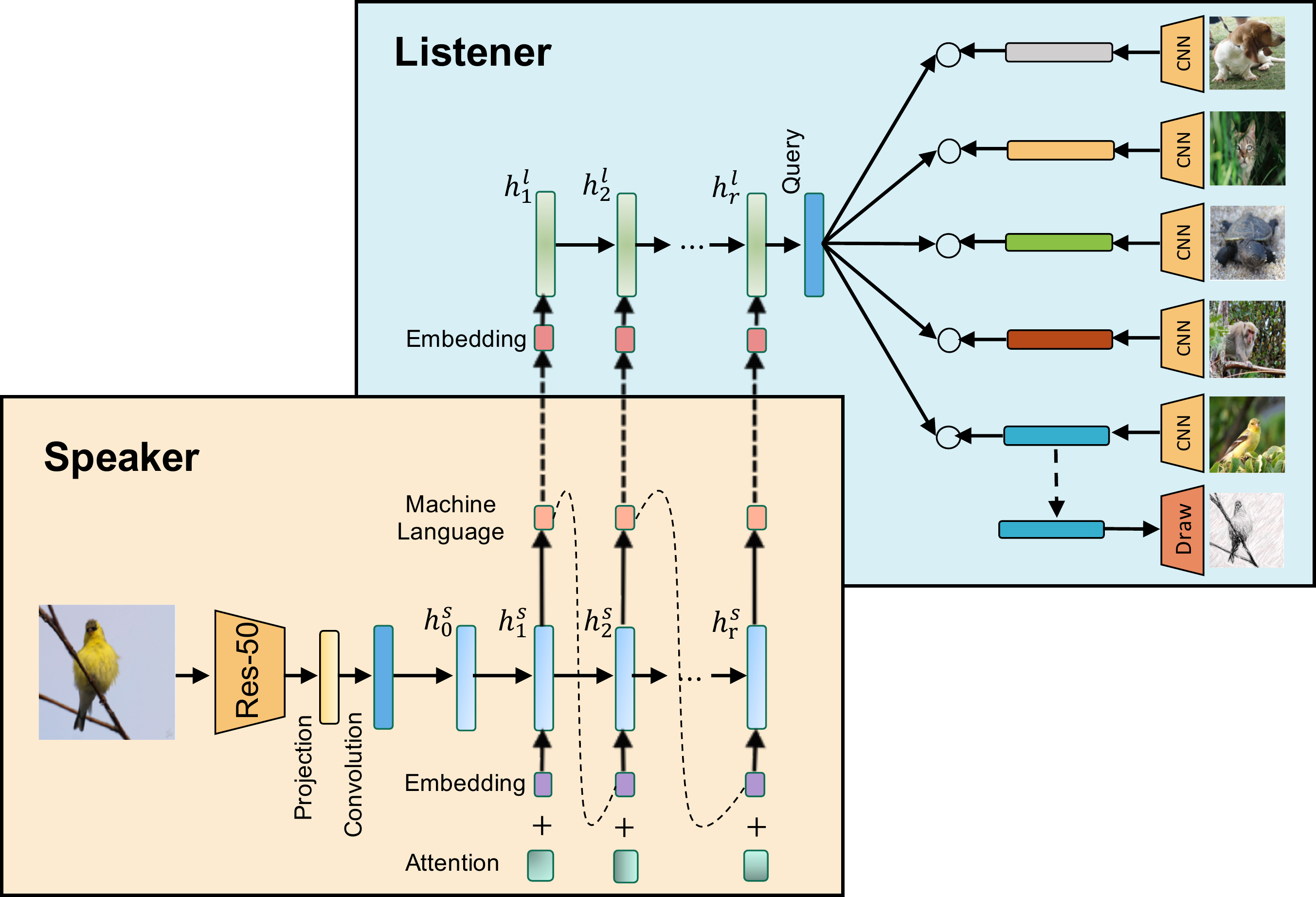}
  \caption{Network Structure of the speaker and listener. We leverage an encoder-decoder architecture. The speaker perceives an image and generates a sequence of symbols that represents machine language. The listener hears the machine language and outputs a query to guess the correct target in a batch. More details can be found in Section 4.}
  \label{fig:fig3}
\end{figure}

We show the detailed network structure in Figure \ref{fig:fig3}. Both speaker $A_s$ and listener $A_l$ are implemented as \textsl{LSTM} networks~\cite{hochreiter1997long}. Given a random image, the speaker will first process this image with a convolutional neural network to extract a feature embedding, based on which a recurrent neural network with \textsl{LSTM} is used to generate a variable-length sequence (the machine language to be learned). After that, the listener will receive this sequence and process it using another recurrent neural network to produce a feature vector which we denote as a query $q$. From the query, the game needs to finish three tasks.
\begin{itemize}
\item \textbf{Guessing}. The listener has not seen the target image before, and the only information he received is the language from the speaker. We ask the listener to guess what the speaker has seen. A batch of images $B$ is randomly selected which contains many irrelevant pictures as well as a target one (seen by the speaker). By using a convolutional neural network to extract features from $B$, we can measure the similarity between $q$ and the features in $B$. A normalization of softmax function can be used on the similarities to produce a probability, then we can calculate a guessing loss  $L_{guess}$ using cross-entropy.
\item \textbf{Drawing}. Guessing is implemented as a selection problem, another task of drawing is further used to promote the semantics of the language. In this process, we ask the listener $A_l$ to draw a picture according to his understanding, and then the reconstruction error between the original image and the drawn one is used to define a loss of $L_{draw}$. This is a generative task that is complementary to the discriminative task of guessing.
\item \textbf{Regularization}. Besides guessing and drawing, we also consider a regularization task. Language should be flexible, and the same object can be described in diverse ways. Therefore, we propose to constrain the description consistency under different sequence lengths of the language. In other words, given an image, the speaker can describe it many times under different sequence lengths, from which the listener will produce different queries. We measure the consistency of these queries as a regularization loss $L_{regularization}$.
\end{itemize}

The guessing loss $L_{guess}$ reflects the instance-level understanding of the picture and expects the model to predict the correct target with a large probability. The drawing loss $L_{draw}$ focuses on the pixel-level comprehension of the picture. Since it is a challenging problem to recover the colorful information of an image based solely on language, the loss here emphasizes the illumination variation (gray-scale). The regularization loss $L_{regularization}$ aims to improve the diversity and consistency of the language. The overall loss function is then the weighted sum of the three, $\lambda_1$ and $\lambda_2$ are hyper-parameters.
\begin{equation}
    L_{total} = L_{guess}+\lambda_1 L_{draw} + \lambda_2 L_{regularization}
\end{equation}

To train this model, we only need some raw images, and data labeling or human language assistance is not required, which is totally a spontaneous process. On the other hand, the flexibility of the language could be enhanced through regularization. Moreover, the design of our games, i.e., coarsely-grained guessing and fine-grained drawing, promotes the semantic information behind the language. All the details of the above methods can be found in Section 4.

\subsection*{2.3 \quad Related Works}
\qquad The process of language generation is very similar to image caption \cite{xu2015show}, which generates language text from visual pictures. The difference is that image caption requires existed language as supervisory signals, and the emergence of a non-exist language should be spontaneous and unsupervised. Recently, some works \cite{gu2018unpaired,liu2018show,feng2019unsupervised} attempt to explore the image caption task from an unsupervised or semi-supervised way. However, they just adjust the process of training and still use pivot language \cite{gu2018unpaired} or additional knowledge from detection and language structure \cite{feng2019unsupervised}. On the contrary, the emergence of language considered in our paper is totally unsupervised and spontaneous. Fascinated by the mystery of human language, researchers mimic the emergence process by multi-agent games \cite{lazaridou2016multi}. Especially in the last five years, simulating language emergence is gaining increasing traction due to the rapid development of deep learning\cite{lecun2015deep} in the studies of language and vision. Given the success of deep learning models in related domains such as image captioning \cite{xu2015show} or machine translation \cite{sutskever2014sequence}, research in this space has recently had something of a resurgence with the introduction of models playing referential games. There are two distinct lines of research. In the first one, emergent language is used as an essential tool for solving tasks in a complex environment. This line of work designs different tasks, aiming to use communication as a means to enhance learning. These tasks can include navigation \cite{das2019tarmac, jaques2019social}, negotiation \cite{cao2018emergent}, translation \cite{lee2018emergent} and so on \cite{mordatch2018emergence}. Another line of work \cite{andreas2016reasoning, havrylov2017emergence, evtimova2018emergent, lazaridou2018emergence, graesser2019emergent,lazaridou2020emergent,mihai2021emergence} focuses on investigating and analyzing the emergence of communication in referential games \cite{skyrms2010signals}, and aligns more closely with our work. There are also theoretical exploration \cite{kottur2017natural,dessi2019focus,kharitonov2019egg,baroni2020rat} on the setting of referential games. However, previous works focus more on the human language simulation rather than a novel language emergence. Our proposed machine language is the first article to generate a non-exist language from the perspective of machines. Different from the previous works that usually use reinforcement learning, we adopt a self-supervised training method. And the setting of the \textsl{SGD (Speak, Guess and Draw)} game is more consistent with our brains, which would reconstruct the visual scene when hearing the language. Although previous work has found linguistic phenomenons during the emergent process, we further analyze the basic natures of the emergent machine language and investigate its advantages compared with continuous features.

\subsection*{2.4 \quad Results and Performance}
\label{sec:ML}
\qquad Our results aim to answer the two questions mentioned above: (1) how does machine language emerge? (2) what makes a sequence of symbols become a language? We first found that machines can generate a language-like sequence during playing the \textsl{SGD} game. Our game setting ensures the requirements of spontaneity. Compared with the communication process in previous works, spontaneity is reflected in the fact that we neither use human language knowledge \cite{havrylov2017emergence, evtimova2018emergent} nor use additional annotations \cite{lazaridou2018emergence}.

\begin{figure}[t]
\begin{minipage}[t]{0.45\linewidth}
\centering
\includegraphics[width=0.9\textwidth]{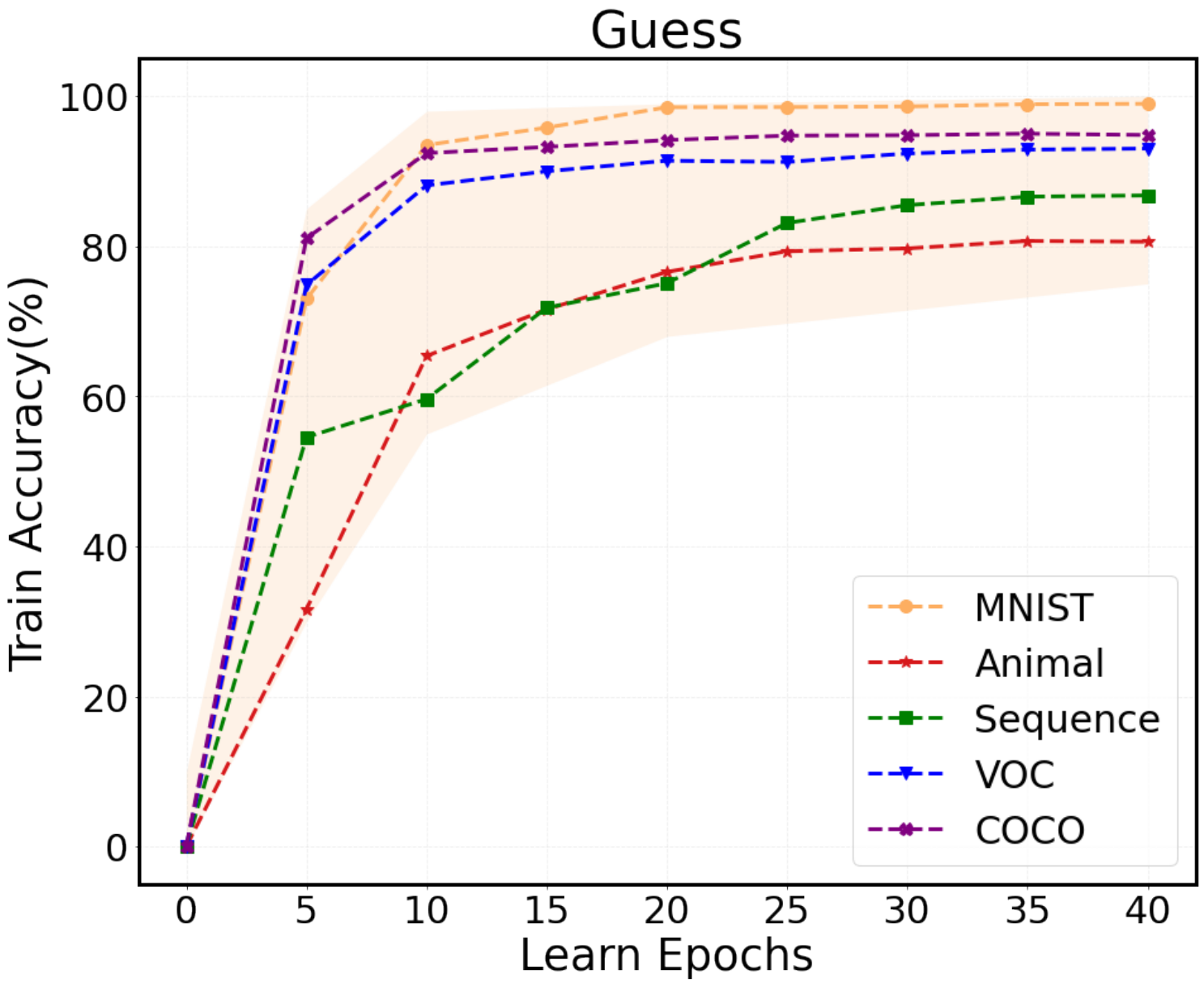}
\caption{\label{fig:fig4} Illustration of the \textbf{training accuracy} through learning epochs. We conduct experiments on five datasets. The accuracy of guessing is increasing with the help of machine language.}
\end{minipage}
\hfill
\begin{minipage}[t]{0.45\linewidth}
\centering
\includegraphics[width=0.9\textwidth]{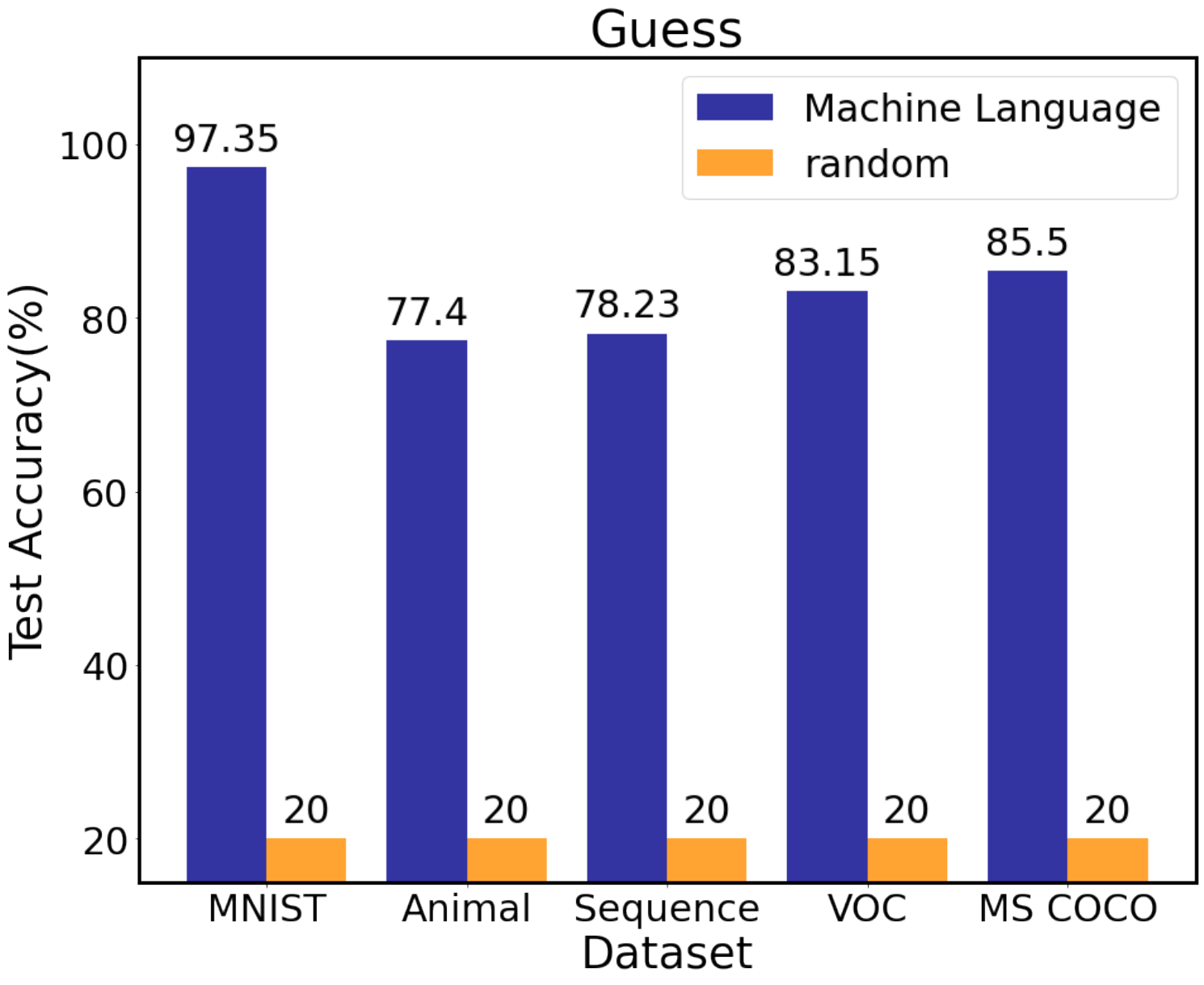}
\caption{\label{fig:fig5} Illustration of the \textbf{test accuracy} compared with the random guess. The results show that machine language has good performance on diverse datasets.}
\end{minipage}
\end{figure}

We conduct experiments on diverse datasets: \textsl{MNIST}, \textsl{Animal}, \textsl{Sequence}, \textsl{VOC}, \textsl{COCO}. The \textsl{MNIST} \cite{lecun1998gradient} contains handwritten digits from $0$ to $9$. The \textsl{Animal} is a subset of \textsl{ImageNet} \cite{deng2009imagenet}, from which we selected $10$ common animals. The \textsl{Sequence} dataset is a variant of \textsl{MNIST} which we combine multiple digits into a sequence. \textsl{Pascal VOC} has $20$ categories, and images are mostly simple scenes and single objects. \textsl{Microsoft COCO} \cite{lin2014microsoft} is a large dataset for common objects in life, and images are mostly complex scenes and have multiple objects. With the help of machine language, agents can play the game well. Figure \ref{fig:fig4} shows the trend of guessing accuracy during training. We set the batch size to $5$, so the accuracy of random guess is 20\%. It can serve as the baseline for our models. However, after only a few epochs of learning, the agent can achieve an accuracy of more than 60\%, far more than the random guess. Figure \ref{fig:fig5} shows the performance of the test set. Our model performs well on all datasets: The accuracy is more than 75\%. Diverse datasets prove the universality and correctness of our approach. Table \ref{tab:table} represents the detailed results on five datasets. Since the target image and sequence length of the language of each time are randomly selected, different initial conditions can lead to fluctuations in performance, so we calculated the average result of $5$ times.

\begin{table}[t]
\centering
\begin{minipage}[t]{0.65\linewidth}
 \caption{\textbf{Guessing Accuracy.} Accuracy of guessing in \textsl{Speak, Guess, and Draw} game on diverse datasets. We set the batch size to $5$, so the accuracy of random guess is 20\%. Each experiments was conducted $5$ times and used average results.}
  
  \begin{tabular}{cccc}
    \toprule
    Method     & Data    & Train Accuracy(\%)& Test Accuracy(\%)\\
    \midrule
    Random & \ & 20.00$\pm$0.00 & 20.00$\pm$0.00 \\
    Machine Language     & MNIST & 99.30$\pm$0.24 & 97.35$\pm$1.07 \\
    Machine Language     & Animal & 88.57$\pm$0.97 & 77.40$\pm$0.88   \\
    Machine Language     & Sequence & 87.65$\pm$0.98 & 78.23$\pm$1.13   \\
    Machine Language    & VOC & 94.35$\pm$1.59 & 83.15$\pm$1.21   \\
    Machine Language     & COCO & 97.04$\pm$0.64 & 85.50$\pm$1.13   \\
    \bottomrule
  \end{tabular}
  \label{tab:table}
  \end{minipage}
\end{table}

\begin{figure}[t]
  \centering
  \includegraphics[scale=0.6]{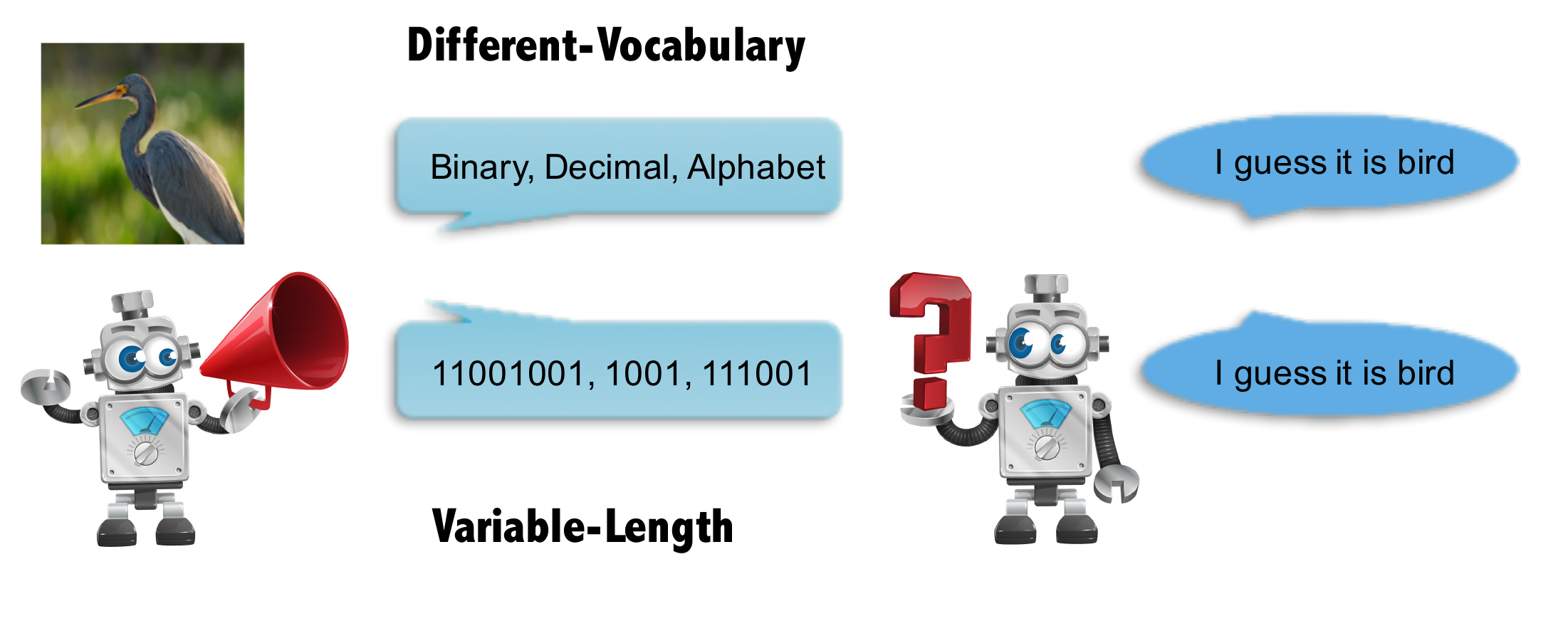}
  \caption{\textbf{Flexibility}. Examples are selected from \textsl{Animal} dataset. We tried different vocabulary sizes, binary (2), decimal (10) and alphabet (26). Experiments show that variable-length sequences can be generated under different vocabularies, and the listener can understand and guess the correct object in the batch.}
  \label{fig:fig6}
\end{figure}

One of the hallmarks of language is flexibility, which contains two aspects. The first requires that the generated language should be a variable-length sequence, but many previous works \cite{lazaridou2016multi,evtimova2018emergent,lazaridou2018emergence} use fixed-length representations. Although works in \cite{havrylov2017emergence,mihai2021emergence} leveraged the idea from image caption and generated a variable-length sequence during communication, they did not match the second aspect of flexibility. Humans can describe a picture using the language of any length. However, the current model only learns the optimal length of a particular picture, and cannot describe it diversely. On the contrary, our method has taken the two aspects of flexibility into account. Even for the same picture, we randomly give a length to the network as conditions, and the network can output a description of the corresponding length. As shown in Figure \ref{fig:fig6}, we show examples on the \textsl{Animal} dataset. Moreover, we found that variable-length descriptions can be generated under different vocabularies. It is similar to human language, i.e., Chinese and English have their own vocabulary to constitute language. Different vocabulary sizes (e.g., binary, decimal, alphabet) could lead to a different language. Although the length and vocabulary of the sequence are different, the listener can understand the language and guess correctly. From the result of the experiment, it proved that the machine language generated in \textsl{SGD} game has the nature of flexibility.

Semantics embodies the function of language, which is the core for distinguishing a language from a meaningless sequence. However, it is very challenging to explore the semantics of machine language, because it is entirely a spontaneous behavior of the machine. There are also some works \cite{havrylov2017emergence,mihai2021emergence} that attempt to measure the semantics captured by an emergent communication protocol. They measure semantics by comparing some indirect indicators, such as mean-rank \cite{mihai2021emergence}. We adopt a more direct method: semantics can be measured by the performance of distinguishing categories.  In \textsl{MNIST} and \textsl{Animal}, we have the category label of the picture. The most straightforward idea is to use machine language as input, the label as output, and learn a mapping by neural networks. From Table \ref{tab:table2}, it turns out that we can successfully construct a mapping from machine language to category. The results in Table \ref{tab:table2} show the top-1 accuracy of successful mapping in the training set and test set. We find that machine language has good performance on both datasets.

\begin{table}[ht]
\begin{minipage}[t]{0.45\linewidth}
 \caption{\textbf{Quantitative Analysis of Semantics.} Classification results on \textsl{MNIST} and \textsl{Animal} dataset. Because these datasets have category information, we can use labels to train a classifier from machine language to category. The input is machine language generated by the speaker, and the output is its category label. From the experiment, we can see that machine language has strong semantics.}
  \centering
  \begin{tabular}{ccc}
    \toprule
    Data    & Train Accuracy(\%) & Test Accuracy(\%) \\
    \midrule
    MNIST  & 99.10 &  97.25    \\
    Animal & 89.80 & 81.87      \\
    \bottomrule
  \end{tabular}
  \label{tab:table2}
\end{minipage}
\begin{minipage}[t]{0.46\linewidth}
 \caption{\textbf{Semantic Analysis on COCO.} We analysis the pattern of machine language on \textsl{COCO} test set. Machine language with specific patterns have common features in images. \textsl{CR} represents the correspondence rate.}
  \centering
  \begin{tabular}{cccc}
    \toprule
    Machine Language     & Features  & Total  & CR(\%)  \\
    \midrule
    N * * * C     & gray-scale & 30 & 0.87 \\
    R * * * I     & person    & 50 & 0.98\\
    B * * * M     & sky       & 46 & 0.93 \\
    G * * * M     & landscape & 67 & 0.91 \\
    T * * * F     & indoor & 90 & 0.97\\
    T * * * G     & food      & 96 & 0.80\\
    Y * * * M     & traffic & 209 & 0.78\\
    \bottomrule
  \end{tabular}
  \label{tab:table3}
  \end{minipage}
\end{table}

The premise of the above analysis is that the dataset has category labels. In real life, there are a lot of images without labels. Here, we use the original pictures of the \textsl{COCO} dataset as an example, and we perform semantic analysis by comparing the contents of pictures behind similar machine languages. We found that the generated machine language usually contains two stages. The symbols at the beginning and end often determine the nature of the object to be described. In Table \ref{tab:table3}, we listed some of the observed pattern in the \textsl{COCO} test set. For example, when the description is $R * * * I$, the corresponding picture is usually a person. When the description is $T * **G$, it is usually a picture of food. The trends summarized from these images may be wrong because it is only from humans' view to understanding a machine's behavior. Therefore, we try to give some features of these images and count the Correspondence Rate (CR). The results in Table \ref{tab:table3} show a high correspondence rate on several examples, except for $Y * **M$. We found that many of the pictures in this pattern are about vehicles and traffic scenes. At the same time, there are lots of wrong pictures containing refrigerators, cabinets, and computers. Machines may think that these things are similar to cars or trains because they all have a similar square shape. Therefore, by quantitative and qualitative analysis on diverse datasets, we conclude that the machine language generated in the cooperation has sufficient semantics.

\begin{figure}[t]
  \centering
  \includegraphics[scale=0.52]{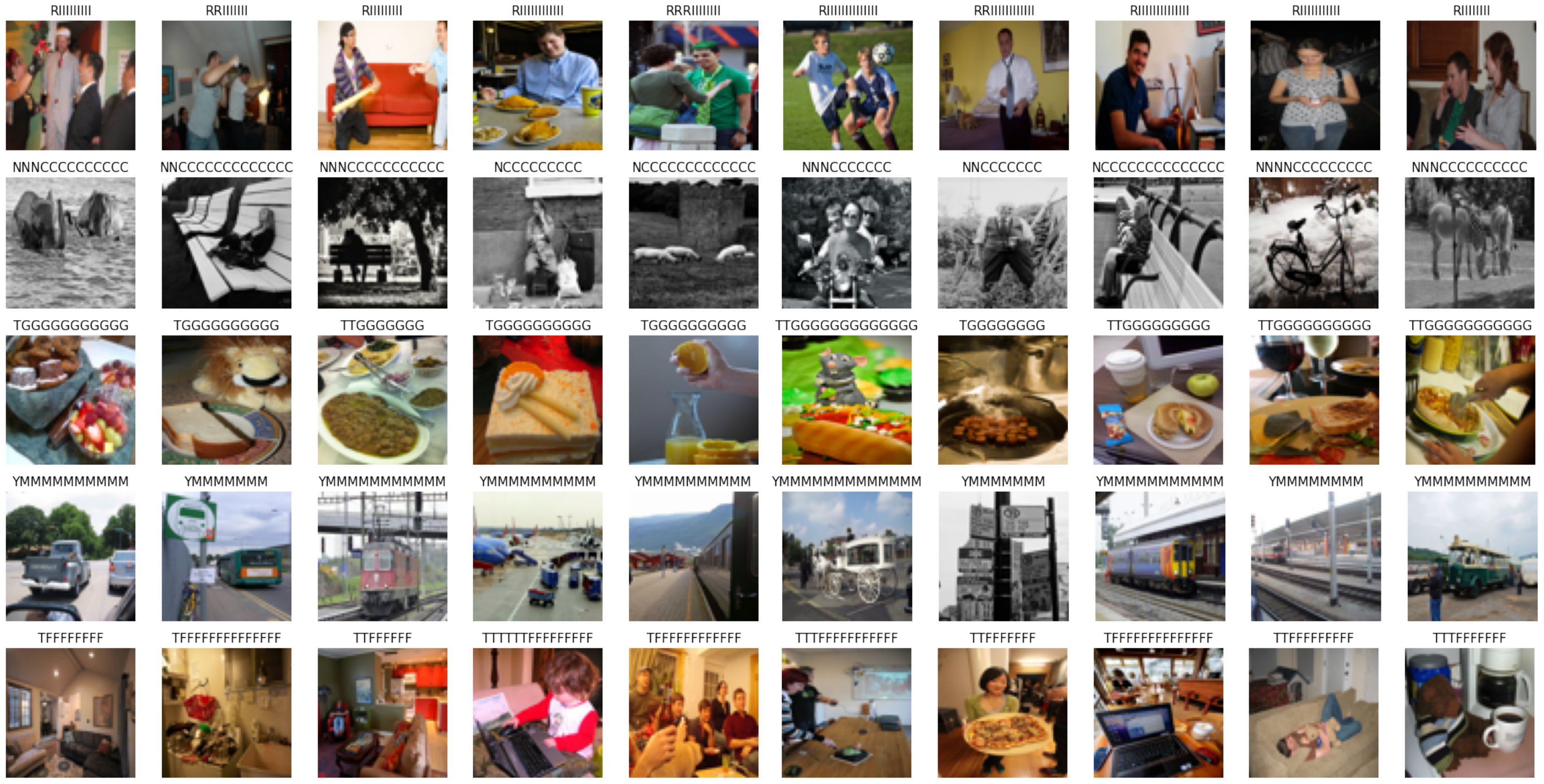}
  \caption{\textbf{Semantic Analysis on \textsl{COCO}}. From the perspective of language, similar languages should have similar visual contents. From top to the bottom, we try to understand these properties from a human perspective. $R * **I$ represents person; $N * **C$ represents gray-scale; $T * **G$ represents food; $Y *** M$ represents traffic transportation; $T *** F$ represents indoor scene.}
  \label{fig:fig9}
\end{figure}

\subsection*{2.5 \quad Discrete Language vs. Continuous Feature}
\qquad The previous section has shown the performance of the emerged machine language. In this section, we will show the advantages of discrete language compared with the continuous feature in the representation and solving of different tasks. Continuous features such as vector, matrix, and tensor are widely used in current deep learning methods. However, learning such a representation is not easy, and often requires a large amount of training data to obtain good generalization. Due to the high-dimensional space and the overfitting problems, continuous representation has several shortcomings including (1) lacking interpretability, (2) poor generalization, and (3) being easy to be attacked. Therefore, we conducted three experiments on these issues to compare the performance between discrete language and continuous feature.

\subsubsection*{Interpretability}
\begin{figure}[t]
  \centering
  \includegraphics[scale=0.6]{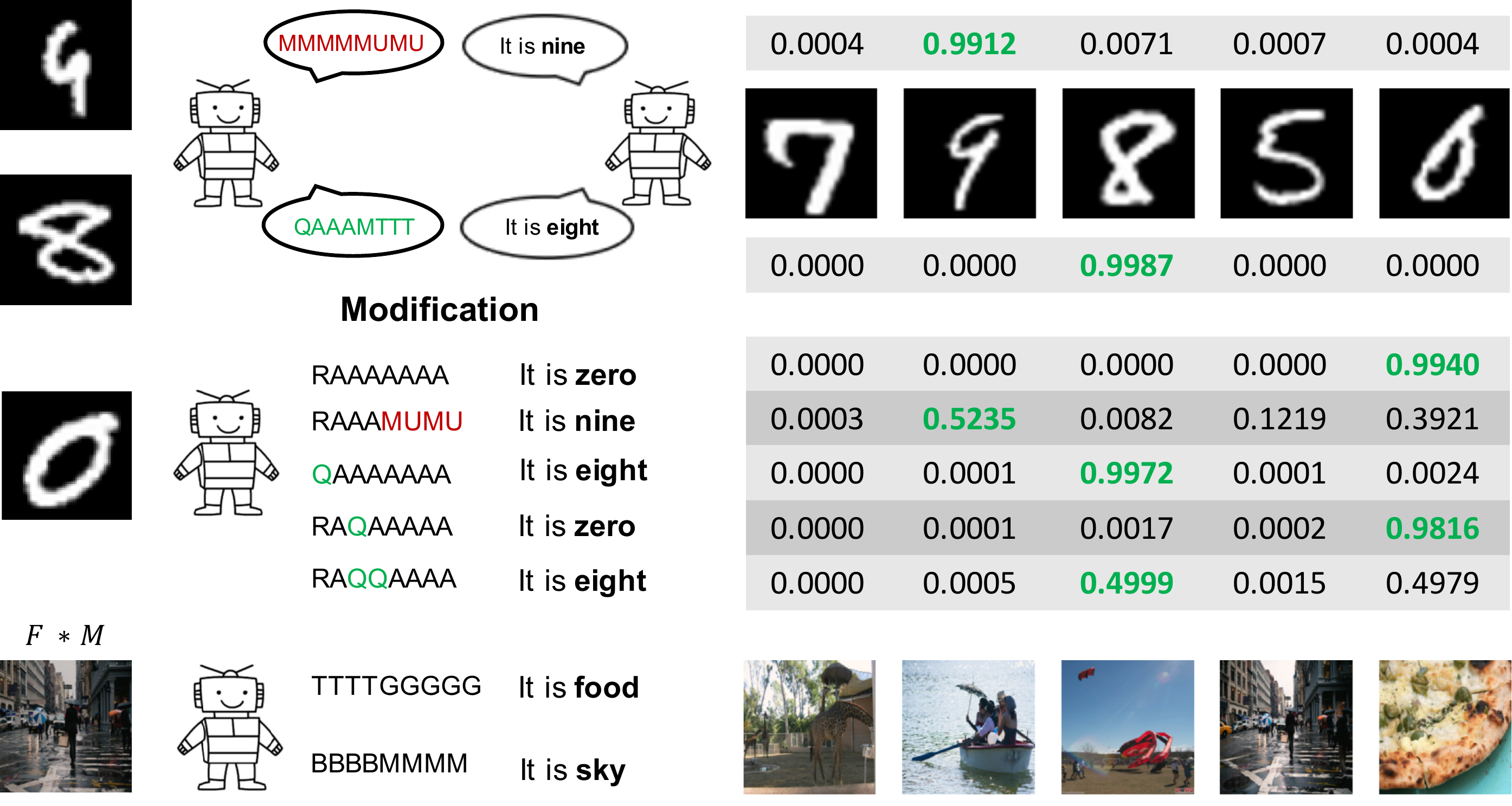}
  \caption{\textbf{Modification of Machine Language}. From top to bottom, the examples of \textsl{MNIST}, modification of \textsl{MNIST} and \textsl{COCO}. The data in the black table shows the confidence probability of listener's prediction. The picture with the highest probability is highlighted in green. Experiments show that we can modify the final decision by changing some symbols directionally.}
  \label{fig:fig8}
\end{figure}

\qquad Language naturally has good interpretability, and we can manipulate and modify the language with a particular purpose to change its semantic meanings. The continuous feature representation learned by deep learning is often high-dimensional and coupled, which is difficult to intuitively understand the content of the representation and modify it directionally. In contrast, the discrete representation of the language is low-dimensional and decoupled, and the learned model can be understood by directional modification of the language. To better understand the communication among machines, we show the decision-making process in Figure \ref{fig:fig8}. From top to bottom, the figure first shows the confidence probability of the listener's judgment when hearing machine language. The speaker outputs $MMMMMUMU$ and the listener believes that it is number $9$ with a probability of $0.99$. The speaker outputs $QAAAMTTT$ and the listener believes that it is number $8$ with a probability of $0.99$. Experiments show that the listener not only guesses correctly but also makes decisions with a high confidence probability. Although the continuous representation also has confidence probability, it is difficult to grasp the key influencing factors due to the complexity of high dimensions. However, we find that machine language has this potential, and we can grasp the key factors through character modification. Specifically, we make some modifications to the machine language. As for digit $0$, the speaker says $RAAAAAAA$ originally. We insert some symbols representing the number $9$ ($MUMU$) into $0$'s machine language ($RAAA$), resulting in $RAAAMUMU$. Surprisingly, we find that the listener's decision would change to $9$. Because some descriptions of the number $0$ are retained, we can see that the option of the number $0$ also has a probability of $0.39$. This discovery tells us that we can know how important certain symbols are, and changing certain symbols can be used to affect the final result. Next, we find that $Q$ is very important for the prediction of the number $8$, so we only modify such a single symbol in the language. As shown in the Figure \ref{fig:fig8}, when the machine language is changed to $QAAAAAAA$, the listener turns to believe the target is $8$. Moreover, the modification of different positions has different effects on the results, which indirectly shows the structure of the language. Furthermore, we also tested it under the \textsl{COCO} dataset of natural scenes. Inspired by the results in the above section, when we modify the machine language to form like $T * G$, the listener believes the target is $food$. When we modify the machine language to form like $B * M$, the listener believes the target is $sky$. These results show that machine language can be manipulated to represent different concepts, and different symbols can be interpreted to reflect different semantic meanings, making the whole decision-making process to be more understandable.  

\subsubsection*{Robustness}
\qquad Robustness is important in practical applications. Although continuous representation methods can achieve excellent performance on different datasets, they often face many challenges in real-life scenarios. Our second experiment focuses on the investigation of robustness between discrete language and continuous feature. To make a fair comparison, we use the same backbone network structure for visual feature extraction, and one model outputs continuous features while another outputs discrete language. Finally, we add a linear layer for classification. This setting is similar to current self-supervised learning, and we can regard the process of language generation as a pretext task. The comparison results are given in Table \ref{tab:table4}. The dataset used in the experiment is \textsl{MNIST}, and we averaged the results for $5$ times. As shown in the first three rows, when there is no noise in the dataset, continuous feature would achieve better performance than discrete language. However, when different noises like Gaussian noise and Salt-Pepper noise are added to the test set, the performance of feature-based representation will significantly drop, while the language-based representation will be more stable and robust. This is easy to understand from the perspective of language, because language is more abstract, and hence more robust to changes in visual details. With more training data, this problem will be alleviated to some extent, however, language still shows consistent advantages over feature. At the same time, we find that Salt-Pepper noise is more destructive to the model than Gaussian noise. In terms of robustness, adversarial examples \cite{szegedy2014intriguing} have attracted much attention in recent years. We also use a black-box attack to generate adversarial examples for the test set. The results in Table \ref{tab:table5} also show that discrete languages are more robust in front of adversarial samples than continuous features.

\subsubsection*{Generalization}
\qquad The last comparison is from the generalization perspective. Continuous feature has good generalization performance in independent and identical distribution (i.i.d.) setting which is proved by many kinds of research. However, we argue that language will have better generalization performance in out-of-distribution (o.o.d) setting due to the compositionality of language. In light of this, we consider the performance of the representation on transferring to new categories. Current supervised learning systems require much more examples than humans and always perform poorly when transferring to new categories. Therefore, we designed an experiment on \textsl{MNIST} to verify the potentials of the language in generalization performance. Here, the numbers from classes $5$-$9$ and $0$-$4$ belong to different data distributions. We only use the data from the first five classes $0$-$4$ for training and evaluate the transfer performance to the new categories of $5$-$9$ in testing. The results are shown in Table \ref{tab:table6}. For known categories, there is little difference in accuracy between discrete language and continuous features. However, for new unknown categories, the ability of language is much stronger than the feature. The reason is that continuous feature-based representation is easier to overfit, and hence it generally performs poorly when generalizing to new categories, and language is more abstract and more holistic, which will generalize better to new concepts.

\begin{table}[t]
\begin{minipage}[t]{0.45\linewidth}
 \caption{\textbf{Robustness on Noise.} Classification results on \textsl{MNIST} when facing noise on the test set. Language representation uses the machine language described by the speaker as the input of the classifier. The feature uses the vector extracted by \textsl{CNN} after average pooling. Both of them are added a linear layer for classification.}
  \centering
  \begin{tabular}{cccc}
    \toprule
    Data     & Noise  & Language(\%)  & Feature(\%)  \\
    \midrule
    1\%      & w/o  & 44.31 & \textbf{66.25}\\
    5\%     & w/o & 84.33  & \textbf{89.54}\\
    10\%     & w/o  & 94.04  & \textbf{97.13}\\
    1\%     & w/ Gaussian    & \textbf{40.74} & 11.30\\
    5\%     & w/ Gaussian  & \textbf{78.27} & 62.79\\
    10\%     & w/ Gaussian  & \textbf{91.36} & 84.11\\
    1\%      & w/ Salt-Pepper    & \textbf{18.19} & 10.19\\
    5\%      & w/ Salt-Pepper    & \textbf{45.08} & 33.12\\
    10\%     & w/ Salt-Pepper    & \textbf{73.69} & 59.37\\
    \bottomrule
  \end{tabular}
  \label{tab:table4}
\end{minipage}
\begin{minipage}[t]{0.45\linewidth}
 \caption{\textbf{Robustness on Adversarial examples.} Classification results on \textsl{MNIST} against adversarial examples.}
  \centering
  \begin{tabular}{ccc}
    \toprule
    Adversarial  & Language(\%)  & Feature(\%)  \\
    \midrule
     w/o & 97.66 & \textbf{99.10}\\
     w/ Adversarial  & \textbf{68.17} & 43.75\\
    \bottomrule
  \end{tabular}
  \label{tab:table5}
\\[12pt]
    \caption{\textbf{Transfer to New Category.} Classification results on \textsl{MNIST} when transferring to new category. The results show that language has better generalization ability in the face of unknown new categories.}
  \centering
  \begin{tabular}{ccc}
    \toprule
    Class      & Language(\%)  & Feature(\%) \\
    \midrule
    0-4      & 99.66 & \textbf{99.77}\\
    5-9      & \textbf{71.72} & 63.50\\
    \bottomrule
  \end{tabular}
  \label{tab:table6}
\end{minipage}
\end{table}

\section*{3 \quad Discussion}
\label{sec:discussion}
\qquad Representation is a core issue in the field of artificial intelligence. In history, artificial intelligence has transitioned from symbolism to connectionism. Symbolic systems have logical meanings, but they are not as powerful as connectionism in learning relations. Connectionism, especially today's deep learning, occupies a dominant position due to its high accuracy on various tasks. The deep neural networks have endowed features with powerful representational capabilities. However, in the face of high-dimensional coupling representation, people can only regard the model as a black box and infer the principle of the model from the changes of input and output. It has also led to the rejection of deep learning models in many risk-averse areas. Increasingly, researchers are arguing that the continuous representation of intelligence maybe not fit the way humans perceive the world. Human language is the most common example in life, which utilized discrete symbols as representation. Human beings compose different symbols to describe the world and endow each discrete symbol with a unique meaning. The composition of the language allows humans to use finite vocabulary to produce infinite meanings and representations. Besides, language is naturally interpretable and has many excellent properties.

It is a very challenging question whether machines could generate language. Because until today, how human language came into being is still a mystery. Recently, many works have used multi-agent cooperation to simulate human communication. However, these works usually used human language as the label in the model training process. We believed the emergence of a non-existent language should be spontaneous. We proposed the \textsl{SGD} game from the perspective of spontaneity, which is simple and more realistic. But what makes a sequence of symbols become a language? Human language has many characteristics, and the emergent machine language here is far from the natural language. Therefore, we draw ideas from language ranging from generation, form, and function, and define three properties for machine language: spontaneity, flexibility, and semantic. Spontaneity means that the generation process of machine language is spontaneous, without external force driving or forcing. Flexibility reflects the formation of language, which is a crucial factor that distinguishes information compression and sparse coding. Semantic carries the functionality of language which should be unambiguous, communicable, and understandable. Our Experiments proved that we indeed generate machine language in \textsl{SGD} games and that the sequence of symbols was spontaneous, flexible, and semantic, which is consistent with our motivation and definition of machine language. We have also shown that machine language has many excellent potentials as a representation of intelligence. In terms of interpretability, symbolic representation has a natural advantage, and by visualizing the confidence probability and manipulating symbols in machine language, we can better understand the decision-making process. In terms of robustness, discrete languages are more robust in the face of noise and adversarial examples, which also provides a new view for the resistance of the model under attack. In terms of generalization, we discussed the ability of language to transfer knowledge to new categories, because language can describe unseen objects through the reorganization of symbols. However, these experiments are only a few initial attempts, and their practical versatility needs further research on more datasets and tasks.

Our model achieved good results on simple datasets, and performance on more complex datasets still needs to improve. Communication in real scenarios is more diverse and complex. Real-life communication is mutual and multi-round, but our model is only one-way information transmission. In the future, we can improve our approach in three aspects. Firstly, the tasks of cooperation can be more complex and diverse. Compared with interpersonal communication in human society, communication in our game is simple, and the simplicity is just for proof. Therefore, the language learned in our model is relatively low-level, carrying only some global semantic information. Complex tasks would promote the generated language to be more advanced. We suppose emergent linguistic phenomena would be more significant to study in more sophisticated settings. Recall the three stages of human language development: (1) emergence, (2) development, (3) evolving. We only explore the first stage of machine language. Can we further create the grammar of machine language? Can the network learn the structure and syntax? Machines can generate more advanced language when combining with more complex tasks. Recently, Geoffrey Hinton \emph{et al.} proposed \textsl{Pix2Seq} \cite{chen2021pix2seq} and cast object detection as a language modeling task. This inspires us to use machine language on the task of object detection. At present, the machine language generated by our game describes information from a global perspective, and the object detection task will pay attention to more local information in the picture, thus lead to fine-grained machine language. Secondly, the network structure used by agents is fundamental, with little or no task-specific tweaking. \textsl{LSTM} network is now adopted in our approach, and we can use models with better performance and more specific designs. We can leverage ideas from Transformer \cite{vaswani2017attention}, which shows great potential in the field of language and vision nowadays. Finally, the communication environment can be more complex. Information exchange can be extended to more agents as well as more tasks to design a multi-agent communication framework for examining high-level linguistic phenomena at the community level.

In summary, we believe that machine language is an exciting and worthwhile direction to study. Artificial intelligence is moving from perceptual intelligence to cognitive intelligence. In a recent Turing lecture \cite{bengio2021deep}, Yoshua Bengio \emph{et al.} proposed the concepts for System $1$ and System $2$ intelligence and believed the future of AI should be system $2$. System $1$ represents current deep learning, which is intuitive, unconscious, and has implicit knowledge, while system $2$ should be logical, sequential, and has explicit knowledge. Our work inherited the above ideas, aiming at reconciling symbolic intelligence with neural networks. This paper is just an initial step, and hopefully, it will shed light on the future development of artificial intelligence research.

\section*{4 \quad Methods}
\label{sec:methods}

\subsection*{Datasets}
\qquad \textbf{MNIST:} The \textsl{MNIST} dataset contains handwritten digits from $0$ to $9$. The original images were $28 \times 28$ pixel gray-scale, and we processed them into $64 \times 64$ images. We use the normal split for training images and test images. As for the training set, we built $5000$ batches of data. Each batch contains $5$ pictures. Note that in the same batch, the digit of each picture is different, and it is randomly sampled from $10$ digits. As for the test set, we built $890$ batches.

\textbf{Animal:} We selected $10$ animals from \textsl{ImageNet} \cite{deng2009imagenet}. The input image is $256 \times 256$ color image. As for the training set, we built $4000$ batches of data. Each batch contains $5$ pictures. Note that in the same batch, the animal of each picture is different, and it is randomly sampled from $10$ animals. As for the test set, we built $1000$ batches.

\textbf{Sequence:} We randomly combine the digits in \textsl{MNIST} dataset. The sequence can be two numbers, three numbers, and four numbers, for example, $23$, $367$, $8907$. The resolution of the combined image is $64 \times 64$. As for the training set, we have made $15000$ images, including $5000$ for $2$ digits, $5000$ for $3$ digits and $5000$ for $4$ digits. As for the test set, we have made $3000$ images, including $1000$ for $2$ digits, $1000$ for $3$ digits and $1000$ for $4$ digits.

\textbf{Pascal VOC:} \textsl{PASCAL Visual Object Classes (VOC)} \cite{everingham2010pascal} is one of most popular datasets in computer vision. We used $33000$ images in \textsl{VOC12} as training set, and used $9900$ images in \textsl{VOC07} as test set.

\textbf{MS COCO:} To explore the experiment in a more general situation, we choose \textsl{MS COCO} \cite{lin2014microsoft} dataset. We only use the original image data without any annotation. The training set and test set are split originally. We used $82000$ images in the training set to build training data and used $40000$ images in the test set to build test data.

\subsection*{Network Structure}
\qquad The overall structure adopts the idea of Auto-Encoder. The encoder can be treated as a speaker, and the decoder as a listener. The speaker encodes the visual feature $v$ into a variable length sequence $M=(m_1,m_2,...,m_r)$ using an encoder $A_{s}(v,m)$. The speaker computes its word $m_t = A_s(v_t,m_{t-1})$ at each time step. In fact, each output should be a distribution of the vocabulary $V$. We use the greedy strategy here in both training and test time, i.e., $m_t = \mathop{\arg\max}_{i} p(m_i,i \in V)$. Network parameters detailed below: the embedding dim of the word $m$ is $128$, the hidden dim in LSTM is $256$, the dropout rate is $0.1$. For the convolutional network, we chose the first $4$ stages of ResNet-50 \cite{he2016deep}. Suppose the picture is $I \in R^{3\times H \times W}$, the visual feature extracted by \textsl{CNN} is $v \in R^{512 \times \frac{H}{16} \times \frac{W}{16}}$. We added a convolution layer to reduce the feature dimension to $128$. The listener $A_l$ uses a bidirectional \textsl{LSTM} network to decode the machine language $M$ heard from the speaker and outputs a query. The model has $2$ layers. The hidden dim is $256$, and the query dim is $128$.

For the classification task, we use a simple structure, only $2$ layers of bidirectional LSTM network. The input of the network is the machine language spoken by the speaker, a sequence of symbols in variable length. The output of the network is the predicted label. The embedding dim of the discrete symbol is $128$, and the hidden dim of LSTM is also $128$. For a machine language $M = (m_1,m_2,...,m_r)$ of length $r$:
$$
    p = \sigma(f(h_r))
$$

$p$ is the predicted category probability, and $\sigma$ is a sigmoid function, and $f(.)$ represents the mapping from hidden dim to category. $h_r$ is the last output of the \textsl{LSTM} model. We use the cross-entropy loss to train the classification network.

\subsection*{Loss Functions}
\qquad The loss function used to train the model consisted of three terms. The first one is a prediction loss function for the \textsl{SGD} game, we defined it as $L_{guess}$. In order to make the predicted probability distribution consistent with the real distribution, we use cross-entropy loss here.
$$
    L_{guess} = -\sum_{x}p(x) \log q(x)
$$

$p(x)$ represents the predicted probability distribution, that is, the probability of predicting a target in a batch. $q(x)$ stands for the real distribution. The second loss is a reconstruction loss function. We expect the agent to restore the information of the real target as much as possible according to the description. It is just like a sub-task to draw the original target, and we defined it as $L_{draw}$. We find that this loss could help the model learn better language descriptions.

$$
    L_{draw} = \left\{
        \begin{array}{ll}
            0.5(x_i-y_i)^{2}, &  |x_i-y_i|<1 \\
            |x_i-y_i|-0.5,   & otherwise
        \end{array}\right.
$$

Our model can only accomplish the painting of the gray-scale image. We used SmoothL1 loss here. $x_i$ represents the gray value of the pixel drawn at position $i$, and $y_i$ represents the real pixel value of the target image. The third loss is a regularization loss, which aims to constrain the description consistency under different lengths. Although the description length is varied, the information obtained by decoding should be the same.
$$
    L_{regularization} = \frac{1}{N} \sum_{i}^{N}{(\bar{q}-q_i)^2}
$$

Suppose we describe the target picture in $N$ different lengths, $q_i$ represents decoded information under a certain description length. $\bar{q}$ represents the average decoding information. The overall loss function is then the weighted sum of the three loss functions:
$$
    L_{total} = L_{guess}+\lambda_1 L_{draw} + \lambda_2 L_{regularization}
$$

The hyper-parameter used in our experiments are shown in the follows:
\textsl{MNIST}: $\lambda_1 = 30.0, \lambda_2 = 10.0$; \textsl{Animal}: $\lambda_1 = 8.0, \lambda_2 = 10.0$; \textsl{Sequence}: $\lambda_1 = 6.0, \lambda_2 = 8.0$; \textsl{VOC}: $\lambda_1 = 6.0, \lambda_2 = 8.0$; \textsl{MS COCO}: $\lambda_1 = 6.0, \lambda_2 = 8.0$.

\subsection*{Training}
\label{sec:train}
\qquad Different from the previous works, we did not use reinforcement learning. Each agent is trained using self-supervised learning. We randomly selected a target from a batch of images. The target image is used as the supervision signal of the network. The batch size was set to $5$, which means each batch has $1$ target and $4$ distracting images.

Model was trained for $40$ epochs using the AdamW-optimizer($\beta_1 = 0.9, \beta_2 =0.999$) with learning rate of $0.0005$. The schedule of the learning rate using the cosine annealing strategy. Here is something to note, the target image that the speaker sees is not the same as the listener sees in the batch. In the \textsl{MNIST} and \textsl{Animal} dataset, both target images belong to the same category, but they have morphological differences. In the \textsl{Sequence},\textsl{VOC} and \textsl{MS COCO}, we change the color and scale of the target image.

\subsection*{Experiment Settings}
\qquad In the emergence of machine language, we use the self-supervised method to train the model. Each time, we randomly select a picture in the batch as the target. The vocabulary size is $26$, and the length range of the language is $8-16$. Length is given randomly in each round of communication.

In the experiment of robustness, the dimension of the feature is $128$, and the length of language is $8-16$. When Gaussian noise is added, the kernel size is $11$, and the sigma is $1.0$ and $2.0$. When adding salt pepper noise, the density is $0.1$. We used \textsl{ZOO} \cite{chen2017zoo} to generate adversarial examples. In the experiment of generalization, the original \textsl{MNIST} data set is divided into two parts: The first part contained the digits from $0$ to $4$ and the second part consisted of the remaining digits from $5$ to $9$. We use the first part to train the model, then fix the model parameters, fine-tune a linear classifier in the second part, and finally test the performance of the model on the new category, from $5$ to $9$.

\bibliography{main}

\end{document}